\documentclass[final]{cvpr}
\usepackage{times}
\usepackage{epsfig}
\usepackage{graphicx}
\usepackage{amsmath}
\usepackage{amssymb}
\usepackage{multirow}
\usepackage{appendix}

\usepackage[pagebackref=true,breaklinks=true,colorlinks,bookmarks=false]{hyperref}

\def\cvprPaperID{8560} % *** Enter the CVPR Paper ID here
\def\confYear{CVPR 2021}

\begin{document}

%%%%%%%%% TITLE
% \title{Zero-Shot Human-Object Interaction and Object Affordance}
\title{Affordance Transfer Learning for Human-Object Interaction Detection}
% \title{Learn to Detect Human Interactions via \\ Combinatorial Affordance Learning}

\author{Zhi Hou$^1$, Baosheng Yu$^1$, Yu Qiao$^{2,3}$, Xiaojiang Peng$^{4}$, Dacheng Tao$^1$ \\
$^1$ School of Computer Science, Faculty of Engineering, The University of Sydney, Australia \\
$^2$ 
% Shenzhen Key Lab of Computer Vision and Pattern Recognition, \\
Shenzhen Institute of Advanced Technology, Chinese Academy of Sciences \\
$^3$ Shanghai AI Laboratory \\
$^4$ Shenzhen Technology University \\
{\tt\small zhou9878@uni.sydney.edu.au, baosheng.yu@sydney.edu.au, yu.qiao@siat.ac.cn,} \\ {\tt\small pengxiaojiang@sztu.edu.cn, dacheng.tao@sydney.edu.au}
% \and
% Yu Qiao \\
% Shenzhen Key Lab of Computer Vision and Pattern Recognition, Shenzhen Institutes of Advanced Technology, Chinese Academy of Sciences \\

% Shenzhen Key Lab of Computer Vision and Pattern Recognition, Shenzhen Institutes of Advanced Technology, Chinese Academy of Sciences

}
\maketitle
% \pagestyle{empty}
% \thispagestyle{empty}

%%%%%%%%% ABSTRACT
\begin{abstract}
Reasoning the human-object interactions (HOI) is essential for deeper scene understanding, while object affordances (or functionalities) are of great importance for human to discover unseen HOIs with novel objects. Inspired by this, we introduce an affordance transfer learning approach to jointly detect HOIs with novel objects and recognize affordances. Specifically, HOI representations can be decoupled into a combination of affordance and object representations, making it possible to compose novel interactions by combining affordance representations and novel object representations from additional images, \ie transferring the affordance to novel objects. With the proposed affordance transfer learning, the model is also capable of inferring the affordances of novel objects from known affordance representations. The proposed method can thus be used to 1) improve the performance of HOI detection, especially for the HOIs with unseen objects; and 2) infer the affordances of novel objects. Experimental results on two datasets, HICO-DET and HOI-COCO (from V-COCO), demonstrate significant improvements over recent state-of-the-art methods for HOI detection and object affordance detection. Code is available at \url{https://github.com/zhihou7/HOI-CL}.
\end{abstract}

%%%%%%%%%%%
\section{Introduction}

Human-object interaction (HOI) detection aims to localize the human and objects in a given image, and recognize the interactions between the human and objects~\cite{chao2018learning}. Considering the combinatorial nature of HOIs, there are always a variety of rare or unseen interactions with novel objects (\eg, ``ride tiger"), which remains a great challenge for the HOI detection model to detect unseen interactions with those novel objects.

%Object affordances are those action possibilities that are perceiveable by an actor\cite{norman2002the, gibson1979The}.

\begin{figure}
\centering
\includegraphics[width=0.9\linewidth]{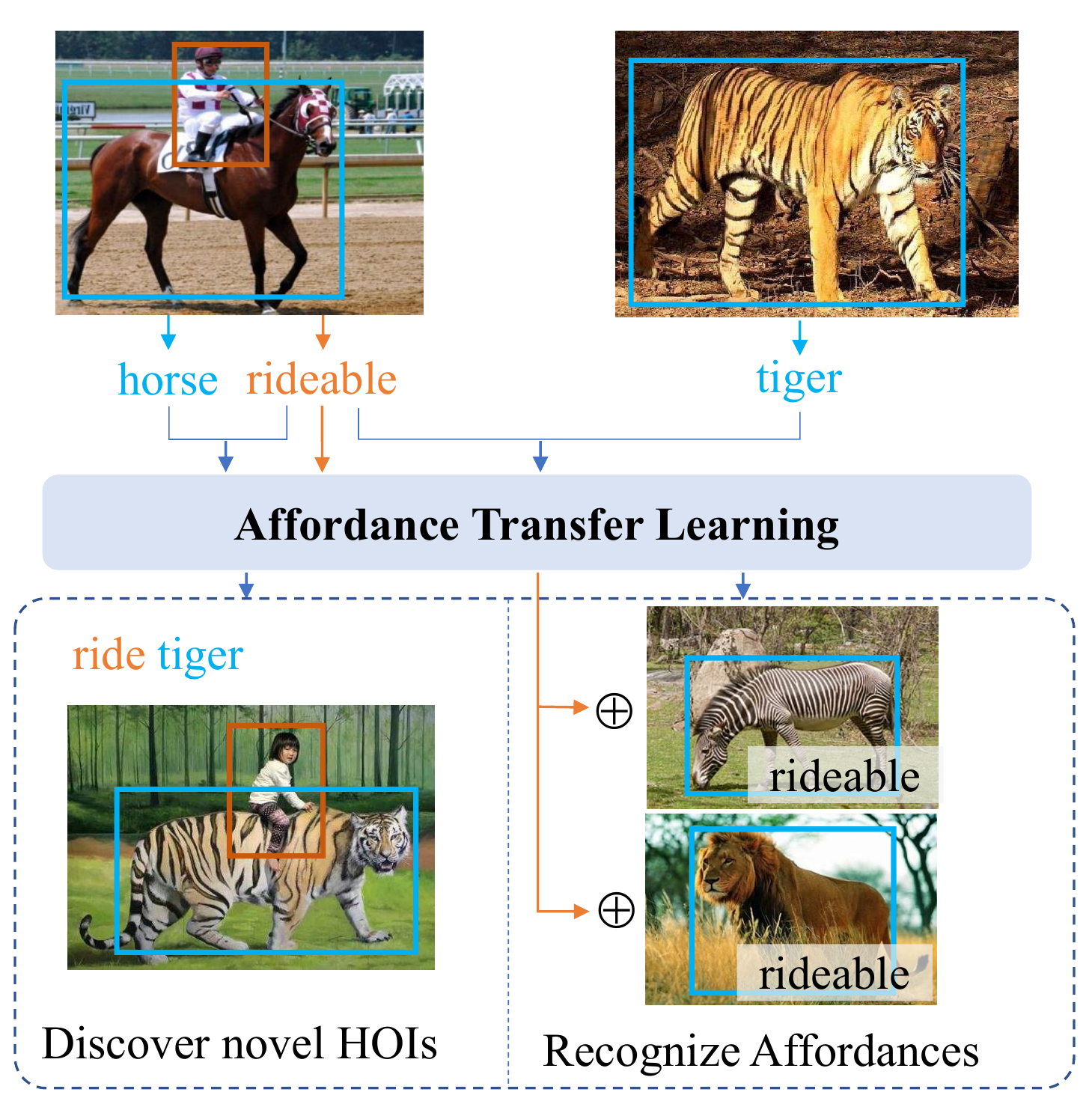}
\caption{An intuitive example to demonstrate affordance transfer learning for jointly exploring human interactions with novel objects (\eg, ``tiger''), and recognizing the affordance of novel objects. The proposed method is able to learn from the unseen interaction samples (\eg, ``ride tiger'') that are composed from affordance representations and novel object representations, which meanwhile transfers the affordance to novel objects and enables the object affordance recognition.}
\label{fig:demo}
\end{figure}

The interactions between the human and object, $\left \langle human, verb, object \right \rangle$, can be captured by either a human-centric (actions) or object-centric (affordance) manner. Specifically, each HOI can be disentangled into a verb and an object, in which the verb also indicates one of the possible affordances (or functionalities) of the object~\cite{gkioxari2018detecting, hassanin2018visual}, \ie what actions can be applied to a particular object \cite{gibson1979The}. Therefore, we are able to jointly learn the affordances of object from the HOI samples, making it possible to compose new HOIs by combining the affordance representations in existing HOIs with novel object representations. Meanwhile, the composition of object representations and the corresponding affordance representations transfers the affordance representation (verb) to novel objects, which we term as affordance transfer learning or ATL. The affordance transfer learning empowers the shared affordance representation learning among different objects, and further facilitates the detection on HOIs with novel objects. For example, with the shared affordance representation (\eg, ``rideable'') between ``tiger'' and ``horse'' as illustrated in Figure~\ref{fig:demo}, we are able to compose new HOIs (\ie, ``ride tiger''), and thus enable the detection of unseen HOIs. The proposed affordance transfer learning framework further generalizes the compositional learning for HOI detection~\cite{hou2020visual} with the ability to detect HOI with additional unseen objects, rather than only the objects from existing HOI samples.

The proposed affordance transfer learning also empowers the HOI detection model to learn object affordance in a weakly supervised manner. Recent HOI detection approaches \cite{gao2018ican, li2018transferable, bansal2019detecting, liao2019ppdm, hou2020visual, gao2020drg} usually fail to explore the possibility of object affordance recognition with the HOI detection model, and previous affordance learning methods \cite{hassanin2018visual} largely ignore transferring shared affordance representations from existing HOIs to novel objects by HOI detection model. By composing new HOI samples from the affordance representations and novel object representations, affordance transfer learning enables the HOI model to distinguish whether a novel object representation can be combined or not with an affordance representation (\ie, verb). We thus recognize object affordances with HOI model as follows: 1) we maintain a feature bank of decoupled affordance representations from the HOI detection dataset; 2) we extract object representations from additional object detection datasets using the same HOI backbone network; and 3) we combine the object representations with all affordance representations in the feature bank as the input of the HOI classifier.
Finally, we are able to obtain a set of HOI predictions, which are further used to infer the object affordances. Overall, the main contribution of this paper can be summarized as follows,
\begin{itemize}
    \item We introduce an affordance transfer learning framework to exploit a broader source of data for HOI detection, especially for human interactions with novel objects. 
    \item We incorporate HOI detection network with decomposed affordances to infer the affordance of novel objects.
    \item The proposed method not only improves recent state-of-the-art HOI detection methods but also facilitates the recognition of object affordance at the same time.
\end{itemize}

%%%%%%%%%%%%%%%%%
\section{Related Works}

\subsection{HOI Understanding}
Human-object interaction (HOI) detection are receiving increasing attention from the community~\cite{chao2018learning}. HOI detection aims to not only detect object and human in an image, but also reason the relationships between human and objects. Since Gputa \etal \cite{gupta2009observing} presented a Human-Object Interaction approach, massive traditional methods \cite{yao2012recognizing, yao2011human} were introduced for HOI recognition using spatial relation \cite{gupta2009observing}, pose \cite{yao2012recognizing}, human part \cite{yao2011human} in the early. Recently, Chao \etal \cite{chao2015hico} introduced a large HOI recognition data HICO \cite{chao2015hico} and a challenging HOI detection dataset HICO-DET \cite{chao2018learning} for HOI understanding. Meanwhile, Gupta \etal \cite{gupta2015visual} introduced the V-COCO dataset, which mainly focuses on the grounding of verbs and their semantic roles.

Currently, there are a large number of HOI detection approaches \cite{gao2018ican, li2018transferable, xu2019learning, wang2019deep, shen2018scaling, bansal2019detecting, peyre2019detecting, wang2020discovering, ulutan2020vsgnet, zhong2020polysemy, kim2020detecting} improving the HOI benchmark \cite{chao2018learning, gupta2015visual, hou2020visual}. According to the target, current methods can be categoried into two-stage HOI detection, which contains common HOI detection \cite{gao2018ican, li2018transferable, wan2019pose, wang2019deep, liuamplifying} and few-and zero-shot HOI detection \cite{shen2018scaling, bansal2019detecting, xu2019interact, hou2020visual, peyre2019detecting, kim2020detecting, wang2020discovering,liu2020consnet}, and one-stage HOI detection \cite{liao2019ppdm, wang2020learning, kim2020uniondet}. Recently, Hou \etal \cite{hou2020visual} propose a visual compositional learning framework to compose novel HOI samples between pair-wise HOI images for low-and zero-shot HOI detection. However, VCL \cite{hou2020visual} can not compose human-novel-object interactions and ignores the possibility of affordance recognition with HOI model. We introduce a novel framework, Affordance Transfer Learning, to transfer the affordance representation to novel objects via composing affordance and novel object representations, and thus enable the detection of HOIs with novel objects.

% One of popular low-and zero-shot methods is factorized method \cite{bansal2019detecting, shen2018scaling, kim2020detecting} that predicts verb and object separately.

\begin{figure*}
\centering
\includegraphics[width=0.9\textwidth]{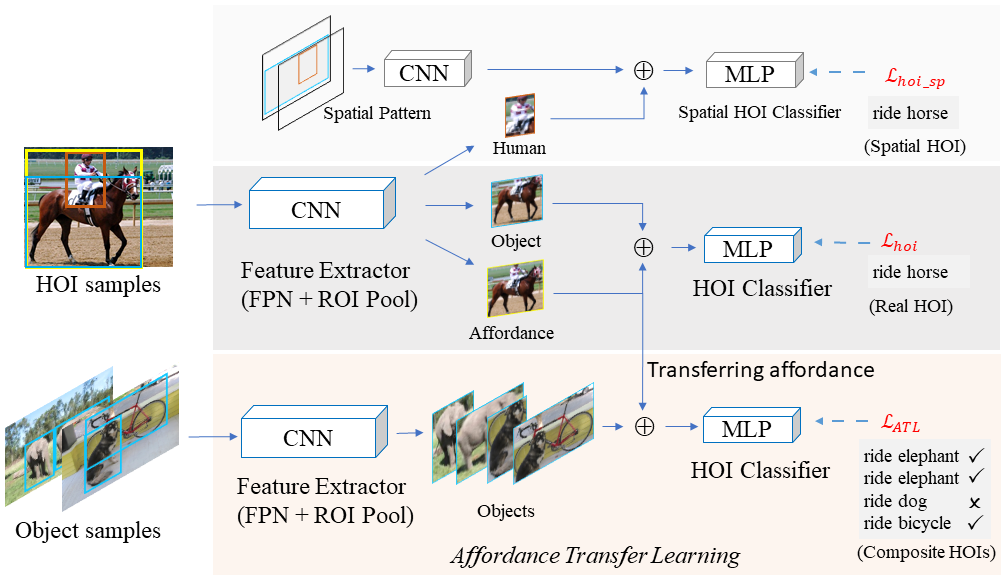}
\caption{An overview of affordance transfer learning or ATL for HOI detection. We first extract the human, object, and affordance features via the ROI-Pooling from the feature pyramids~\cite{ren2015faster}, respectively. Meanwhile, we also extract new object features from an additional object datasets using the same backbone network. After that, we concatenate the affordance and the object features (from HOI datasets) as the real HOIs. We also compose new HOIs using the affordance features and the object features extracted from additional object datasets, which transfer the affordance to novel objects. Both the composite HOIs and real HOIs share the same HOI classifier. In addition, human features and spatial pattern features are combined to construct the spatial HOI branch.}
\label{fig:pipeline}
\end{figure*}

\subsection{Object Affordance}
% For object perception, one popular idea is that human perceive objects by color, shape, size, etc, while Gibson \cite{gibson1979The} proposed an important idea about object perception that human perceive objects by looking at their affordances.
James J. Gibson defined the word affordance in \cite{gibson1979The}. Object affordances are those action possibilities that are perceiveable by an actor \cite{norman2002the, gibson1979The, hassanin2018visual}, that is also the possibilities of Human-Object Interactions. In the early, Kjellstr \etal \cite{kjellstrom2011visual} investigated learning the affordances of objects from human demonstration. Yao \& Li \etal \cite{Yao2013Discovering} presented a weak supervised approach to discover object funtionalities from HOI data in the environment where the person is interacting with musical instruments. Fouhey \etal \cite{Fouhey2014PeopleWH} introduced an approach to estimate functional surfaces by observing human actions. Recently, Fang \etal \cite{fang2018Demo2Vec} introduces Demo2Vec to learn interaction region and action label from online video. Differently, we demonstrate to learn a shared affordance representation with HOI model, and recognize object affordance via classifying the composite HOIs of shared affordance representations from existing HOIs and the target object representation.

%Object affordances are those action possibilities that are perceiveable by an actor\cite{norman2002the, gibson1979The}.

%%%%%%%%%%%%%%%%%
\section{Method}

In this section, we first give an overview of the proposed method, and then introduce the affordance transfer learning for HOI detection. Lastly, we describe the recognition of object affordances with the proposed HOI detection model.

\subsection{Overview}

The motivation of affordance transfer learning is to transfer the affordance to novel objects for exploring unseen HOIs, \ie,  combining the affordance representations and novel object representations. The affordance transfer learning meanwhile enables HOI network to recognize object affordance. Similar to~\cite{gao2018ican, xu2019learning, gupta2019no, ulutan2020vsgnet}, we utilize the popular two-stage HOI detection framework: 1) we first detect the objects in a given image using a common object detector (\eg, Faster-R-CNN~\cite{ren2015faster}); and 2) we then construct HOIs from object and affordance representations to perform HOI classification. The main framework of our affordance transfer learning is illustrated in Figure~\ref{fig:pipeline}, which consists of three branches, spatial HOI, real HOI, and composite HOI. Inspired by~\cite{gao2018ican, li2018transferable, hou2020visual}, we utilize the spatial HOI branch to further improve the HOI detection performance. Specifically, the spatial pattern representation consists of two $64\times64$ binary feature maps to indicate human and object relative positions, \ie, the pixels within the human (or object) bounding box are assigned the value 1. Both real HOIs and composite HOIs are constructed from object/affordance features and share the same HOI classifier~\cite{sadeghi2011recognition}, while the difference between them is that the objects in the composite HOIs are extracted from additional object image datasets. Furthermore, by transferring the shared affordance representations extracted from HOI samples to novel objects, we are also able to use the HOI detection model to recognize the affordance of novel objects.

\subsection{Affordance Transfer Learning}

The proposed affordance transfer learning first composes novel HOI samples between object representations from additional object images (\eg, images from COCO dataset \cite{lin2014microsoft}) and decoupled affordance representation as illustrated in Figure~\ref{fig:pipeline}, and then generalize the affordance representation to novel objects via jointly optimizing the network with the composite HOIs. With the additional objects, the affordance transfer learning effectively decouples the affordance representation from the scenes, and then enables the composition of affordance and novel objects to recognize the affordance of novel objects. In this subsection, we introduce how to efficiently compose new HOIs and remove invalid HOIs for affordance transfer learning.

{\bf Efficient HOI Composition.} The label assignment for the composite HOIs can be integrated from the verb label (or the affordance label) and the object label. The object label $\widetilde{l}_o$ is provided by the object datasets, and we obtain the verb label $l_v$ by decoupling the HOI label. Similar to \cite{hou2020visual}, we decouple the HOI label space into a verb-HOI co-occurrence matrix $\mathbf{A}_v \in R^{N_v \times C}$ a the object-HOI co-occurrence matrix $\mathbf{A}_o \in R^{N_o \times C}$, where $N_v$, $N_o$, and $C$ indicate the numbers of verbs, objects and HOI categories, respectively. Here, both $\mathbf{A}_v$ and $\mathbf{A}_o$ are binary matrix. Given the one-hot HOI label $y \in R^C$, we then obtain the verb label $l_v$ from $y \mathbf{A}_v^T$. To compose a new HOI by the object $\widetilde{l}_o$ and verb $l_v$, we assign the label to the composite HOI as follows,
\begin{equation}
\label{eq:compose}
    \bar{y} = (\widetilde{l}_o \mathbf{A}_o) \& (l_v \mathbf{A}_v),
\end{equation}
where $\&$ indicates the element-wise ``and'' logical operation. Both $\widetilde{l}_o \mathbf{A}_o$ and $l_v \mathbf{A}_v$ are binary vectors (\ie HOI labels), which represent all possible HOIs corresponding to $\widetilde{l}_o$ and $l_v$, respectively. Therefore, the intersection between $\widetilde{l}_o \mathbf{A}_o$ and $l_v \mathbf{A}_v$ indicates the label of the verb-object pair $\left\langle l_v, \widetilde{l}_o \right\rangle$, \ie, the label of the composite HOI.

{\bf Invalid HOI Elimination.} With additional object datasets, we can compose a large number of new types of HOI samples by combining object and affordance features. Considering a variety of object categories, there is nevertheless some invalid HOIs (\eg, ``ride orange''), \ie , the invalid HOIs are out the space of ground truth HOI labels. Furthermore, the same verb might have different meanings in different scenes \cite{Gella2017Disambiguating, zhong2020polysemy}, while the verbs in current HOI dataset (\eg, HICO-DET) mainly represents action (affordance) and are usually not ambiguous~\cite{Gella2017Disambiguating}. Meanwhile, a variety of recent HOI detection methods do not distinguish the same affordance among different HOIs~\cite{shen2018scaling, bansal2019detecting, gupta2019no, kim2020detecting, xu2019interact, peyre2019detecting}. To this end, we also equally treat the same affordance from different HOIs for the evaluation of the transfer affordance learning. Following \cite{hou2020visual}, we simply remove those HOIs, which is out of the HOI label space (\eg, ``ride dog'' in HICO-DET) as illustrated in the right part of Figure~\ref{fig:pipeline}. In addition, the one-hot labels of those invalid HOIs are all zeros according to Equation~\ref{eq:compose}. That is, we can easily remove those composite HOIs according to the one-hot labels.

\subsection{Object Affordance Recognition}

\begin{figure}
    \centering
    \includegraphics[width=0.45\textwidth]{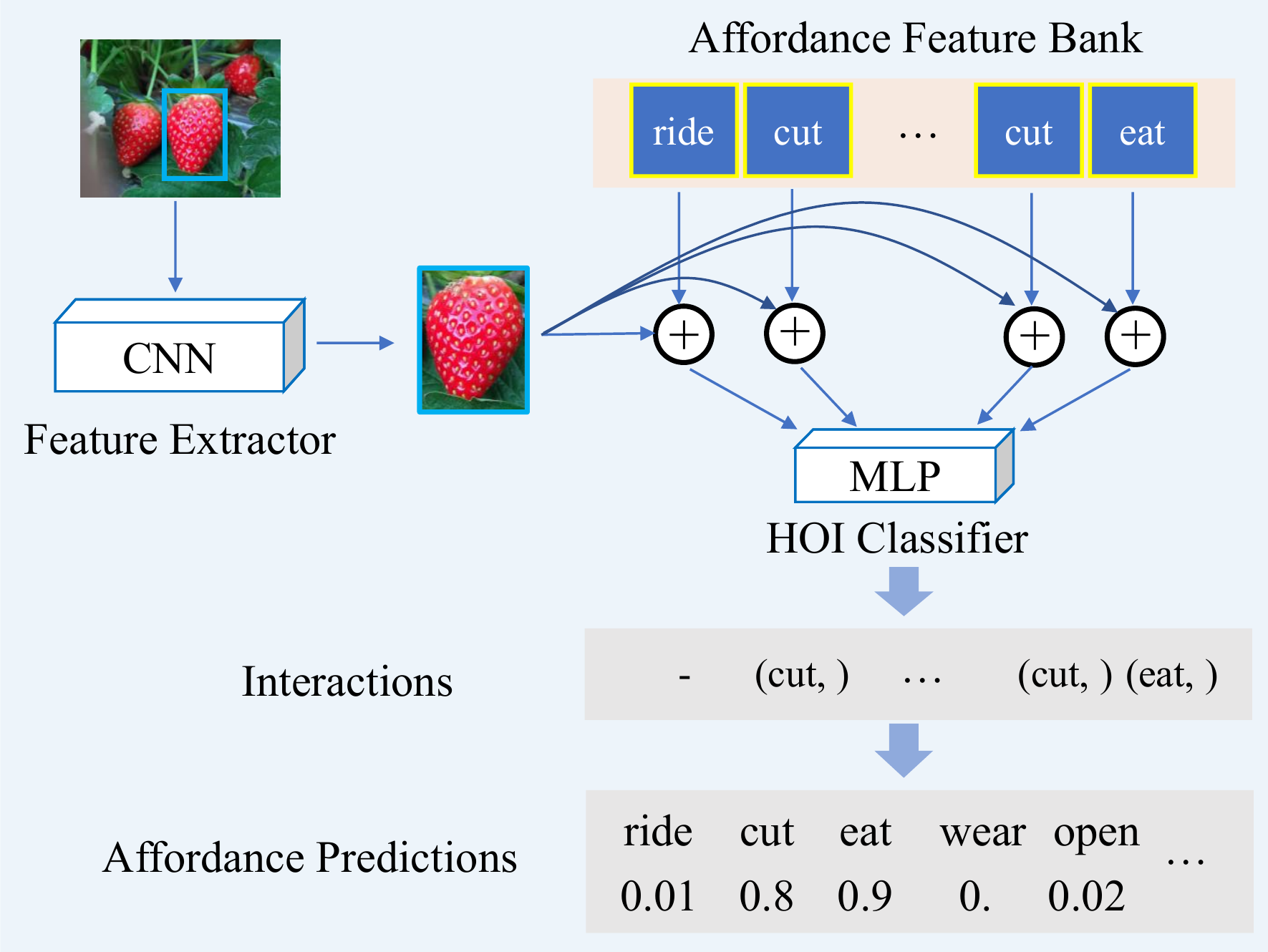}
    \caption{An illustration of object affordance recognition with HOI network. Here, we use verb to represent affordance. We first construct an affordance feature bank from the decoupled affordance representations. For any object (\eg strawberry), we extract the object feature by the Feature Extractor according to bounding box. Then, the object feature is combined with all affordances in the bank to input into HOI classifier for obtaining predicted interactions. The interactions are further converted into affordances (\eg eatable).}
    \label{fig:obj_func}
\end{figure}

In this subsection, we introduce how to infer the object affordance during the testing phase. Considering that we jointly optimize the decoupled components (\ie, object features and affordance features from object and HOI images) in HOI samples and novel object samples with affordance transfer learning, the proposed method thus is able to distinguish whether a novel object is combinable or not with a specific affordance (\ie, valid HOIs). Therefore, we design a simple yet effective object affordance recognition method using the HOI detection model. Specifically, we first build an affordance feature bank as follows.

{\bf Affordance Feature Bank}. We construct the affordance feature bank from HOI datasets (\eg HICO-DET and HOI-COCO). In order to reduce storage space and computation, we randomly choose a maximun of $M$ instances for each affordance in HICO-DET. In our experiment, $M$ is 100. Then, we extract the features of those affordances to construct an off-the-shelf affordance feature bank.

Given an object feature extracted from the object image, we combine it with all affordances in the feature bank to obtain a set of HOIs. As illustrated in Figure~\ref{fig:obj_func}, we obtain all HOI predictions from the HOI classifier. After that, we are able to convert all HOI predictions to affordance predictions according to the HOI-verb co-occurrence matrix $\mathbf{A}_v$. Specifically, we remove the predicted affordances whose label is not the same as the corresponding affordance labels in the feature bank. As a result, we obtain a list of affordances with many repeated elements. Let $F_i$ denotes the frequency (count) of the affordance $i$ and $S_i$ indicates the number of affordance (or verb) $i$ in the feature bank, we evaluate the probability of the affordance $i$ as $\frac{F_i}{S_i}$.

\subsection{Optimization and Inference}

During the training stage, we train the proposed method with an unique loss $L_{ATL}$ for transferring the affordance to novel objects. Meanwhile, similar to \cite{gao2018ican}, we also incorporate spatial pattern loss $L_{sp}$ to optimize the learning of the HOI spatial pattern representation. In addition, an HOI loss $L_{hoi}$ is used for HOI classification. Lastly, the overall training loss function is defined as follows,
\begin{equation}
    \mathcal{L} = \mathcal{L}_{hoi\_sp} + \lambda_1 \mathcal{L}_{hoi} +  \lambda_2 \mathcal{L}_{ATL},
\end{equation}
where $\lambda_1$ and $\lambda_2$ are two hyper-parameters to balance different losses. Both the feature extractors and the HOI detection modules are jointly trained in an end-to-end manner. $\mathcal{L}_{hoi\_sp}$, $\mathcal{L}_{hoi}$, $\mathcal{L}_{ATL}$ are binary cross entropy losses.

During the testing stage, affordance transfer learning module is not necessary. Similar to \cite{gao2018ican, li2018transferable, hou2020visual}, we predict the final HOIs with spatial HOI predictions and verb-object HOI predictions. Formally, given a human-object bounding box pair ($b_h$, $b_o$), we predict the score $S^c_{h,o}$ as $s_h \cdot s_o \cdot s^c_{hoi} \cdot s^c_{sp}$ for each HOI category $c \in 1,...,C$, where $C$ denotes the total number of possible HOI types, $s_h$ and $s_o$ are the human and object detection scores respectively, $s^c_{sp}$ represents the spatial HOI prediction score and $s^c_{hoi}$ is the verb-object HOI prediction score.

%%%%%%%%%%%%%%%%%%%
\section{Experiments}

We conduct a number of experiments to evaluate the our method for HOI detection using two HOI datasets: HICO-DET \cite{chao2018learning} and HOI-COCO (built from V-COCO \cite{gupta2015visual}). Furthermore, we also evaluate the HOI detection model with affordance transfer learning for object affordance recognition on COCO dataset \cite{lin2014microsoft}, HICO-DET dataset \cite{chao2018learning}, and COCO classes and non-COCO classes in Object365 \cite{Shao2020Objects365}.

\subsection{Datasets and Evaluation Metrics}

{\bf HICO-DET~\cite{chao2018learning}} dataset consists of 38,118 images in the training set and 9,658 test images over 600 types of interactions (80 object categories in COCO dataset and 117 unique verbs) with over 90,000 HOI instances.

{\bf HOI-COCO} is built from the V-COCO dataset~\cite{gupta2015visual}, which contains 10,346 images with 16,199 person instances. Each annotated person in V-COCO has binary labels for 26 different actions. V-COCO mainly focuses
on verb recognition, and has limited object categories (only
two). Thus we construct a new benchmark HOI-COCO for the evaluation of verb-object pairs as follows. We use 21 actions from all 26 actions in V-COCO (\ie, five non-interaction actions, ``walk", ``run, ``smile", ``stand" and ``point ", are removed). As a result, we build HOI-COCO benchmark with 222 HOI categories over 21 verbs and 80 objects. Meanwhile, we use the same train/val split in V-COCO for HOI-COCO. Similar to HICO-DET \cite{chao2018learning}, we evaluate the performance on HOI-COCO under three different settings: Full (222 types), Rare (97 types), and NonRare (115 types). The HOI type in Rare category contains less than 10 training instances, and the distribution of HOI categories is long-tailed.

{\bf COCO \cite{lin2014microsoft}} dataset is a widely-used benchmark for common object detection with 80 different object classes. Considering that both HICO-DET \cite{chao2018learning} and HOI-COCO consist of the same object label sets to COCO, we thus directly incorporate the COCO dataset as the additional object dataset in our experiments.

{\bf Object365 \cite{Shao2020Objects365}} is a recently proposed large-scale common object detection dataset with 365 object categories. The domain of Object365 is different from COCO \cite{lin2014microsoft}. In detail, we select objects that are labeled as COCO classes from Object365 validation dataset to evaluate the affordance recognition of objects on new domain. Meanwhile, we choose 12 new types of objects and label manually the affordance of those objects according to the HICO-DET and HOI-COCO, respectively. Those objects are used to evaluate affordance recognition on new types of objects. See more details in supplementary materials.

{\bf Evaluation Metrics}.
We follow the standard evaluation metric \cite{gao2018ican, xu2019learning} and report mean average precision for HICO-DET dataset \cite{chao2018learning} and HOI-COCO. A prediction is a true positive only when the detected human and object bounding boxes have IoUs larger than 0.5 with reference to ground truth, and the HOI category is accurately predicted. Object affordance recognition is a multiple label classification problem (\ie an object usually has multiple affordances). Thus, we compare Precision, Recall and F1-Score for evaluating object affordance recognition.

\subsection{Implementation Details}
For HICO-DET, similar to recent methods~\cite{bansal2019detecting, liao2019ppdm, wang2020learning, hou2020visual}, we use the object detector fine-tuned on HICO-DET, \ie the detector provided in \cite{hou2020visual}. For HOI-COCO, we directly use the object detector pre-trained on COCO. Besides, all HOI classifiers consist of two fully-connected layers with 1024 hidden units. To compare with recent methods on HICO-DET, we use two object images in each mini-batch. On HOI-COCO, we only use one object image for evaluation. Besides, we also use an auxiliary verb loss \cite{hou2021fcl} to improve our baseline. During training, following \cite{gao2018ican, li2018transferable, hou2020visual}, we augment the ground truth boxes via random crop and random shift. During inference, we keep human and objects with the score larger than 0.3 and 0.1 on HICO-DET respectively. Following \cite{hou2020visual}, we set $\lambda_1 = 2$, $\lambda_2 = 0.5$ on HICO-DET, and $\lambda_1= 0.5$, $\lambda_2 = 0.5$ on HOI-COCO, respectively. To prevent composite interactions from dominating the training of the model, we keep the number of composite interactions not more than the number of objects in each mini-batch by randomly sampling composite HOIs. We train the model for 1.2M iterations on HICO-DET dataset and 300K iterations on HOI-COCO with an initial learning rare of 0.01. For object affordance recognition, we use the actions of each HOI dataset as affordances and remove the ``no interaction'' categories on HICO-DET dataset. We keep the object affordance predictions if the affordance score is large than 0.5. All experiments are conducted on a single Tesla V100 GPU with TensorFlow~\cite{abadi2016tensorflow}.

\subsection{HOI Detection}

{\bf HICO-DET}. We report the performance on three different settings: Full (600 categories), Rare (138 categories) and NonRare (462 categories) in ``Default" and ``Known" modes on HICO-DET. As shown in Table~\ref{table:sota_hico}, the proposed method outperforms recent state-of-the-art methods among all categories. Furthermore, with better object detection results provided in \cite{gao2020drg}, the performance of ATL dramatically increases to {\bf 28.53\%}. Meanwhile, we find ATL is more effective on Rare category. Specifically, when using the objects from the training set of HICO-DET, the proposed method is similar to VCL \cite{hou2020visual} as shown in Table~\ref{table:sota_hico}. ATL also improves the baseline effectively based on One-Stage method. Here, the baseline is the model without compositional learning. Details of One-Stage method is provided in supplementary materials.

\setlength{\tabcolsep}{4pt}
\begin{table}[tp]
\caption{Comparison to recent state-of-the-art methods with fine-tuned detector on HICO-DET dataset~\cite{chao2018learning}. The content in brackets indicates the source of the object images. The last two rows are one-stage HOI detection results.}
\label{table:sota_hico}
\centering
% \begin{center}
\resizebox{0.95\linewidth}{!}{
\begin{tabular}{@{}lcccccc@{}}
\hline\hline
\multirow{2}{*}{Method} &
\multicolumn{3}{c}{Default}&\multicolumn{3}{c}{Known Object}\cr\cline{2-7}
% \cmidrule(lr){2-4} \cmidrule(lr){5-7}
&Full&Rare&NonRare&Full&Rare&NonRare\cr

\hline\hline

FG~\cite{bansal2019detecting} & 21.96 & 16.43 & 23.62 & - & - & - \\
IP-Net \cite{wang2020learning} & 19.56 & 12.79 & 21.58 & 22.05 & 15.77 & 23.92 \\
PPDM \cite{liao2019ppdm} &  21.73 & 13.78 &24.10 &24.58 &16.65 &26.84 \\
% Kim \etal \cite{kim2020uniondet} & 17.58 & 11.72 & 19.33 & 19.76 & 14.68 & 21.27 \\
VCL \cite{hou2020visual} & 23.63 & 17.21 & 25.55 & 25.98 & 19.12 & 28.03 \\
DRG \cite{gao2020drg} & 24.53 & 19.47 & 26.04 & 27.98 & 23.11 & 29.43 \\

% VCL (reproduced) \cite{hou2020visual} & 23.87 & 17.41 & 25.80 & 26.11 & 19.47 & 28.10 \\
%   22.05 15.77 23.92
%  Ours     22.05 15.77 23.92

\hline
% Baseline &  23.44 & 16.80 & 25.43  & 25.27 & 18.18 & 27.39 \\
ATL (HICO-DET)  & 23.67  & 17.64 &  25.47 & 26.01 & 19.60 & 27.93\\
% Ours (HICO-DET)  & 23.63  & 18.05 &  25.29 \\
% 0.2601  0.6208    0.1960  0.6011     0.2793  0.6267     0.1231  0.2812 0.4655  0.2598

% 173092  0.2363  0.6207    0.1805  0.6006     0.2529  0.6267     0.0927  0.2584 0.4539  0.2359
ATL (COCO) & {\bf 24.50} & {\bf 18.53} & {\bf 26.28} & {\bf 27.23} & {\bf 21.27} & {\bf 29.00} \\
% Ours (HICO-DET, COCO) \\

\hline
% VCL$$ 0.2831  0.8267    0.2123  0.8185     0.3042  0.8291     0.1534  0.3030 0.4833  0.2827     0.3062  0.8267    0.2325  0.8185     0.3282  0.8291     0.1889  0.3242 0.4872  0.3059
% 0.2732  0.8278    0.2096  0.8229     0.2922  0.8292
% ATL (HICO-DET) $^{DRG}$ & 27.32 & 20.96 & 29.22 & 29.92 & 23.61 & 31.81 \\
ATL (HICO-DET) $^{DRG}$ & 27.68 & 20.31 & 29.89 & 30.05 & 22.40 & 32.34 \\
% 450000  0.2768  0.8277    0.2031  0.8229     0.2989  0.8292     0.1632  0.2943 0.4456  0.2766     0.3005  0.8277    0.2240  0.8229     0.3234  0.8292     0.1989  0.3162 0.4539  0.3003
% 129819  0.2732  0.8278    0.2096  0.8229     0.2922  0.8292     0.1533  0.2916 0.4360  0.2729     0.2992  0.8278    0.2361  0.8229     0.3181  0.8292     0.1876  0.3164 0.4449  0.2990     drg31

ATL (COCO) $^{DRG}$ &  {\bf 28.53}  &  {\bf 21.64 }  & {\bf 30.59} &  {\bf  31.18 }  & {\bf 24.15}   & {\bf 33.29 } \\

\hline\hline

Baseline (One-Stage)& 22.77 & 16.54 & 24.63 & 26.31 & 21.60 & 27.72  \\
ATL (One-Stage) & {\bf 23.81} & {\bf 17.43}  & {\bf 25.72}  & {\bf 27.38} & {\bf 22.09} & {\bf 28.96} \\
\hline\hline
\end{tabular}
}
\end{table}

{\bf HOI-COCO}. We find the proposed method has similar performance to VCL when using HOI-COCO as the source of object images in Table~\ref{table:sota_coco}. Here, we evaluate the performance of VCL on HOI-COCO dataset using the official code from~\cite{hou2020visual}. When using the COCO object dataset, the proposed method significantly improves the performance, especially on Rare categories, \eg, over {\bf 1.5\%} than VCL and {\bf 2.9\%} than the baseline, respectively. Meanwhile, the proposed method also gives a larger improvement than baseline in NonRare category comparing with VCL, suggesting that ATL also increases the diversity of HOIs via composing new samples. Furthermore, when using both HICO-DET and COCO to provide object images, we further improve the performance to {\bf 25.29\%}.

%Particularly, the improvement of affordance transfer learning for HOI detection on HICO-DET dataset is worse than that on HOI-COCO dataset. We think there are two reasons. One the one hand, HOI-COCO has only 8537 training instances, while HICO-DET has 93041 training instances. Therefore, HOI-COCO dataset is more lacking in training data. One the other hand, HOI-COCO is selected from COCO dataset and thus the two datasets have similar domain. However, HICO-DET is apparently different from COCO \cite{lin2014microsoft}. Therefore, the object images from COCO datasets can not help HOI detection on HICO-DET as much as HOI detection on HOI-COCO.

% limit the same number of COCO datasets

\begin{table}[tp]
\caption{Comparison to recent state-of-the-art methods on HOI-COCO dataset.}
\label{table:sota_coco}
\centering
\small
% \begin{center}
\resizebox{0.9\linewidth}{!}{
\begin{tabular}{@{}lcccc@{}}
\hline\hline
Method & object data & Full & Rare & NonRare\cr

\hline\hline

% iCAN \cite{gao2018ican}
Baseline & - & 22.86 &  6.87 &  35.27 \\
VCL \cite{hou2020visual} & HOI-COCO & 23.53 & 8.29 & 35.36 \\
%   22.05 15.77 23.92
%  Ours     22.05 15.77 23.92
% Ours &  & - & 23.53 & 8.29 & 35.36 \\
% 240000 23.40 23.40 8.01 35.34

ATL & HOI-COCO & 23.40 & 8.01 & 35.34 \\
% Ours & HICO-DET & 24.21 & 9.52 & 35.61\\
% Ours & COCO & 25.40 & 10.31 & 37.12\\
% 24.22 7.97 36.82
ATL & COCO & 24.84  & 9.79  &36.51 \\

ATL & COCO, HICO-DET &  {\bf 25.29} & {\bf 9.85} & {\bf 37.27} \\

\hline\hline
\end{tabular}}
\end{table}

\setlength{\tabcolsep}{4pt}
\begin{table}[tp]
\centering
%\begin{center}
\caption{Comparison of Zero Shot Detection results of our proposed method. UC means unseen composition HOI detection. NO means novel object HOI detection. * means we only use the boxes of the detection results. Here, the baseline means we do not use affordance transfer learning (\ie without $L_{ATL}$).
  }
\label{tab:zero_shot1}
\small
% \resizebox{\linewidth}{!}{
\begin{tabular}{@{}lcccc@{}}
\hline\hline
Method & Type & Unseen & Seen & Full \cr
\hline\hline

Shen \etal \cite{shen2018scaling} & UC & 5.62 & - & 6.26 \\
FG \cite{bansal2019detecting}  & UC & 10.93 & 12.60 & 12.26 \\
\hline
VCL \cite{hou2020visual} (rare first) & UC & 10.06 & 24.28 & 21.43 \\
% ATL (rare first) & UC & 9.18 & {\bf 24.40} & 21.36  \\
ATL (rare first) & UC & 9.18 & {\bf 24.67} & {\bf 21.57} \\
\hline
VCL \cite{hou2020visual} (non-rare first) & UC & 16.22 & 18.52 & 18.06 \\
ATL (non-rare first) & UC & {\bf 18.25} & {\bf 18.78} & {\bf 18.67} \\

\hline\hline
FG \cite{bansal2019detecting} &  NO & 11.22 &  14.36 & 13.84  \\

% Baseline & UO & 12.86 & 20.77 & 19.45  \\
Baseline & NO & 12.84 & 20.63 & 19.33 \\
ATL (HICO-DET) & NO & 11.35 & 20.96 & 19.36 \\
ATL (COCO) &  NO & {\bf 15.11} &  {\bf 21.54}  & {\bf 20.47} \\

% ATL (HICO-DET)
% Baseline & UO & 12.51 & 21.63 & 20.11 \\
% ATL (HICO-DET) & 11.06 & 21.54 & 19.80 \\
% ATL (COCO) & UO & {14.31} & {\bf 22.24} & {\bf 20.92} \\
\hline
Baseline* & NO & 0.00 & 14.13 & 11.77 \\
ATL (HICO-DET)* & NO & 0.00 & 13.67 & 11.39 \\
ATL (COCO)* &NO & {\bf 5.05} & {\bf 14.69} & {\bf 13.08} \\
% \hline\hline
% Functional Gen \cite{bansal2019detecting} &  UO & 11.22 &  14.36 & 13.84  \\
% Baseline (rare first)  & UO & 12.86 & 20.77 & 19.45  \\
% Ours (rare first) &  UO & {\bf 15.54} &  20.74  & {\bf 19.87} \\

\hline\hline
\end{tabular}
\end{table}

\begin{table*}[tp]
\caption{Comparison of object affordance recognition with HOI network among different datasets (based on Mean average Precision). Val2017 is the validation 2017 of COCO \cite{lin2014microsoft}. Subset of Object365 is the validation of Object365 \cite{Shao2020Objects365} with only COCO labels. Novel classes are selected from Object365 with non-COCO labels. Object means what object dataset we use. ATL$^{ZS}$ means novel object zero-shot HOI detection model in Table 3 on HICO-DET. For ATL$^{ZS}$, we show the results of the 12 classes of novel objects in Val2017, Subset of Object365 and HICO-DET.}
\label{table:func_obj}
\centering
\small
% \begin{center}
\begin{tabular}{@{}lccccc|ccc|ccc|ccc@{}}
\hline\hline
\multirow{2}{*}{Method} & \multirow{2}{*}{HOI Data} & \multirow{2}{*}{Object}
& \multicolumn{3}{c}{Val2017 of COCO}&\multicolumn{3}{c}{Subset of Object365} &\multicolumn{3}{c}{HICO-DET} & \multicolumn{3}{c}{Novel classes}\cr\cline{4-15}
&&&Rec&Prec&F1&Rec&Prec&F1&Rec&Prec&F1&Rec&Prec&F1\\
\hline\hline
Baseline & HOI & - & 28.62 & 32.34 & 27.08 & 21.75 & 22.20 &19.83 & 36.64 &49.83& 37.67
  & 12.39 & 8.63 & 9.62 \\
% VCL \cite{hou2020visual} & 21.96 & 19.56 & 12.79 & 21.58 & 22.05 & 15.77 & 23.92  & 23.92 & 23.92\\ 9.62
% FCL \cite{hou2021fcl} & HOI & - &  29.97 &  56.66  &  36.79 &
% 22.32 &  45.26  & 27.90 & 38.62 &  74.27  &  48.15 & 14.99   & 24.86  & 17.45 \\

VCL \cite{hou2020visual} &  HOI  & HOI & 76.93 & 71.79 & 72.15 & 68.60 & 67.52&  65.82 & 87.98  &82.59& 83.84 & 54.75 & 35.85 &  40.43 \\

% Ours & 21.96 & 21.10 & 14.46 & 23.09 & - & - & -  & - & -\\
% Ours & {\bf 86.55} &   {\bf89.52 } &  {\bf 88.01} &  {\bf 77.71} &  {\bf 85.60} &  {\bf 81.47 }&  {\bf 42.17 } &  {\bf 39.97 }&  {\bf 41.04 }\\

ATL & HOI  & HOI &80.71 & 72.79 & 74.44 & 71.76 & 67.34 & 67.13  & 90.29 & 83.21 & 85.30 &  {\bf58.73} &  37.75&   42.75 \\
% Ours ( V-COCO) & 72.79 &  80.71 & 76.54 &67.34 & 71.76 & 69.48 & 37.77 & 58.74 & 45.98 \\

% Ours &  HOI & COCO & {\bf 92.13}  & {\bf 85.89}  & {\bf 87.65}  & {\bf 85.15}  & {\bf 80.64}  & {\bf 82.83}  & {\bf 93.35} & {\bf 90.77} & {\bf 92.04} & 53.65  &  40.94  & 43.57 \\

ATL &  HOI & COCO & {\bf90.94} &{\bf87.33}&{\bf87.65} &{\bf82.95}& {\bf82.13}&  {\bf80.80}& {\bf93.35}&  {\bf90.77}&   {\bf91.02}&  53.65 &  {\bf40.94} &  {\bf43.57} \\

% Ours (Both) & 92.32 & 84.97 & 88.49 & 86.43 & 83.80 &  85.09 & & & & 49.43 &  38.26  & 43.13 \\
% iCAN_R_union_multi_semi_ml5_l05_t5_def2_aug5_3_new_VCOCO_test_both_CL_21

\hline

Baseline &HICO&- &8.11  & 29.21   &11.81  & 6.77    & 26.1  & 9.97   & 8.11  &  29.21   &15.55  & 8.12 &15.87&8.78 \\

% FCL \cite{hou2020visual}

% VCL \cite{hou2020visual} & 21.96 & 19.56 & 12.79 & 21.58 & 22.05 & 15.77 & 23.92  & 23.92 & 23.92\\
% FCL \cite{hou2021fcl} & HICO & - 
VCL \cite{hou2020visual} &HICO & HICO & 9.63& 43.62&  14.89& 10.66& 38.69 & 15.77  & 10.76  & 53.54 & 16.82 &  7.81 & 22.63 &   11.02  \\

% Ours & 21.96 & 21.10 & 14.46 & 23.09 & - & - & -  & - & -\\
% Ours & {\bf 86.55} &   {\bf89.52 } &  {\bf 88.01} &  {\bf 77.71} &  {\bf 85.60} &  {\bf 81.47 }&  {\bf 42.17 } &  {\bf 39.97 }&  {\bf 41.04 }\\

% ATL  &HICO & HICO& 14.01 & 46.45&  20.24 &17.71& 50.92 &24.61 &15.54&  52.25&  22.54&  {\bf 3.92}& {\bf 8.46}&  {\bf 5.17 } \\
ATL  &HICO & HICO& 14.01 & 46.45&  20.24 &17.71& 50.92 &24.61 &15.54&  52.25&  22.54&  {\bf 12.78} &{\bf 28.8}&{\bf 16.78} \\

ATL &HICO & COCO& {\bf 33.69}  & {\bf 79.54} & {\bf 44.32} & {\bf 28.25} & {\bf 63.56} & {\bf 35.24} & {\bf 30.27}&  {\bf 73.53}&  {\bf 40.31}& 12.41  &  14.56  &  12.86 \\

\hline

ATL$^{ZS}$   &HICO & HICO& 4.28& 22.96&  6.98& 3.54& 19.35 &5.8& 6.02& 32.22&  9.93&  5.02&11.63& 6.79 \\

ATL$^{ZS}$ &HICO & COCO & {\bf19.41}& {\bf66.70}& {\bf29.01}&  {\bf15.57}&{\bf  55.58}&{\bf 23.49}&  {\bf19.36}&{\bf 67.55}&{\bf 28.81} & {\bf 14.00}  &  {\bf 28.60}  &  {\bf 18.07} \\

% Ours (Both) & 92.32 & 84.97 & 88.49 & 86.43 & 83.80 &  85.09 & & & & 49.43 &  38.26  & 43.13 \\

\hline\hline
\end{tabular}

\end{table*}

\subsection{Zero-Shot HOI detection}
The proposed affordance transfer learning enables the detection of HOIs with novel objects due to the mechanism of composing HOI samples of unseen classes. Therefore, we evaluate the proposed method for zero-shot HOI detection on HICO-DET \cite{chao2018learning}. We report the performance on two settings: 1) Unseen Composition and 2) Novel Object. Specifically, Unseen Composition means there are unseen HOIs in the test but the verbs and objects of the unseen HOIs exist in training data, while the objects of unseen HOIs in novel object HOI detection do not exist in training data. For compositional zero-shot learning, we follow \cite{hou2020visual} to evaluate on rare-first unseen HOIs (firstly select tail HOIs in HICO-DET as unseen data) and non-rare first unseen HOIs (firstly select head HOIs in HICO-DET as unseen data). We evaluate zero-shot HOI detection on three categories: Unseen (120 categories), Seen (480 categories) and Full (600 categories). For novel object HOI detection, similar to \cite{bansal2019detecting}, we choose 100 unseen categories (includes 12 unseen objects) and 500 seen categories. We choose the object detector provided in \cite{hou2020visual} to compare fairly with \cite{hou2020visual}.

{\bf Compositional Zero-Shot HOI Detection}. In Table~\ref{tab:zero_shot1}, we find our approach effectively improves the non-rare first zero-shot HOI detection. Meanwhile, our approach achieves better result on seen category in rare first zero-shot HOI detection. Particularly, the affordances in tail part of HOIs are usually rare, the composite samples of tail HOIs with additional objects are much less than that of head HOIs. Therefore, our approach achieves even worse result on unseen category.

{\bf Novel Object HOI Detection}. Table~\ref{tab:zero_shot1} demonstrates that transferring affordance representation to novel objects effectively facilitates the detection of unseen HOIs with novel objects. Here we use the network without affordance transfer learning as our baseline. We find using HICO-DET (remove HOIs with unseen objects) as object images even degrades the performance on unseen categories compared to the baseline because we compose massive seen HOI samples but not unseen HOI samples with HICO-DET. Besides, similar to \cite{bansal2019detecting}, we use an generic object detector to enable HOI detection with novel object, which provides a strong baseline. While we only use the boxes of the detector (not use the object label predicted by detector), the performances of baseline and ATL (HICO-DET) on unseen category decrease to 0. However, ATL (COCO) still achieves 5.05\% on unseen category.

\subsection{Object Affordance Recognition}

Table~\ref{table:func_obj} shows ATL significantly improves the baseline by {\bf over 40 \%} in F1 score among all datasets with COCO categories, and by {\bf over 30\%} on novel object classes on HOI-COCO. On HICO-DET, ATL improves the baseline by {\bf nearly 10\%} among all categories in datasets with COCO categories, and by {\bf around 5\%} on novel object classes. Those experiments indicate that the affordance transfer learning via composing novel HOIs effectively disentangles the affordance and object representations from the scenes and endows the HOI network with the ability of affordance recognition. Noticeably, ATL$^{ZS}$ (COCO), that composing interactions from affordances and novel objects, largely improves the baseline model ATL$^{ZS}$ (HICO) on the 12 novel classes among all evaluation datasets.

%Furthermore, ATL can also be applied to recognize the affordance of novel objects and still achieves dramatic improvement.

With the object images from the HOI dataset, our method is similar to VCL \cite{hou2020visual} on HOI-COCO dataset, because both two methods compose HOI samples between two images. On HICO-DET, the proposed method has a better performance than VCL \cite{hou2020visual}, while the two methods have similar performance on HOI detection, as shown in Table~\ref{table:sota_hico}. We find Recall, Precision and F1-Score are sensitive to the threshold. Thus, we further illustrates the Mean average Precision results of all models in Supplementary Materials.

\subsection{Ablation Studies}

%We ablate the effect of the number of object images on HOI detection, the object detector on the proposed method.

\begin{table}[tp]
\caption{Illustration of the number of object images in each batch on HICO-DET dataset.}
\label{table:obj_images}
\centering
\small
% \begin{center}
%\resizebox{0.9\linewidth}{!}{
\begin{tabular}{@{}cccccc@{}}
\hline\hline
\#Images & Full & Rare & NonRare\cr
\hline\hline
% iCAN \cite{gao2018ican}
% 1 object image & 22.86 &  6.87 &  35.27 \\
% 2 object images  & 23.53 & 8.29 & 35.36 \\
%   22.05 15.77 23.92
%  Ours     22.05 15.77 23.92
% 0 & 23.44 & 16.80 & 25.43 \\
1 & 24.07& 18.17& 25.83\\
2 & {\bf 24.50} & {\bf 18.53} & {\bf 26.28}\\
3 & 24.19 &  17.33 &    26.24 \\

% 4  object images & 24.21 & 9.52 & 35.61\\
\hline\hline
\end{tabular}
\end{table}

%
% \begin{table}[tp]
% \caption{Illustration of the number of object images in each batch on HICO-DET dataset.}
% \label{table:obj_images}
% \centering
% \small
% % \begin{center}
% %\resizebox{0.9\linewidth}{!}{
% \begin{tabular}{@{}ccccccccc@{}}
% \hline\hline
% \#Images & Full & Rare & NonRare & COCO & 365 & HICO & Novel\cr
% \hline\hline
% % iCAN \cite{gao2018ican}
% % 1 object image & 22.86 &  6.87 &  35.27 \\
% % 2 object images  & 23.53 & 8.29 & 35.36 \\
% %   22.05 15.77 23.92
% %  Ours     22.05 15.77 23.92
% % 0 & 23.44 & 16.80 & 25.43 \\
% 1 & 24.07& 18.17& 25.83 & 24.73 & 17.45 &  18.66 & 2.33\\
% 2 & {\bf 24.50} & {\bf 18.53} & {\bf 26.28}& {\bf 44.32} & {\bf 35.24} & {\bf 40.31} & 4.41\\
% 3 & 24.19 &  17.33 &    26.24 & 38.12 & 28.25 & 20.94 & {\bf 4.55} \\
%
%
% % 4  object images & 24.21 & 9.52 & 35.61\\
% \hline\hline
% \end{tabular}
% \end{table}

{\bf The number of object images in each batch}. Table~\ref{table:obj_images} shows ATL achieves best performance with 2 object images. We think more object images increase the diversity of object features and balance the object distribution. However, too much object images also hampers the performance.

\begin{table}[tp]
\caption{Illustration of the effect of different object detectors on HOI detection in HICO-DET. Fine-tuned detector is provided in \cite{hou2020visual}. GT means ground truth boxes. The last column is the detection mAP on HICO-DET test dataset.}

\label{table:obj_detector}
\centering
\small
% \begin{center}
\resizebox{0.9\linewidth}{!}{
\begin{tabular}{@{}cccccc@{}}
\hline\hline
Model & Detector & Full & Rare & NonRare & mAP\cr
\hline\hline
Baseline & COCO & {\bf 21.07} & {\bf 16.79} & {\bf 22.35} & 20.82\\

% VCL & COCO & 20.94 & 16.43 & 22.29 & 20.82 \\
% ATL (HICO) &COCO & \\

ATL & COCO & 20.08 & 15.57 & 21.43 & 20.82 \\
\hline
% VCL & Fine-tuned & 23.63 & 17.21 & 25.55 & 30.79\\

Baseline & Fine-tuned & 23.44 & 16.80 & 25.43 & 30.79\\
ATL & Fine-tuned & {\bf 24.50 } &  {\bf18.53}  & {\bf26.28}  & 30.79 \\
\hline
%
% VCL 0.4309 &  0.3256 &  0.4624 \\
Baseline & GT & 43.32 & 33.84 & 46.15 & 100\\
ATL & GT & {\bf44.27}  & {\bf35.52}  & {\bf46.89}  & 100\\

% 90000  0.2344  0.5605    0.1680  0.5534     0.2543  0.5626     0.0890  0.2568 0.2527  0.5605    0.1818  0.5534     0.2739  0.5626     0.1063  0.2752

\hline\hline

\end{tabular}}
\end{table}

% We think it might be because we extract object features according ground truth boxes and augmented boxes of ground truth boxes.

{\bf Object detector}. Due to the domain shift between HICO-DET and COCO, COCO detector usually achieves worse result. we thus use the same fine-tuned object detector as \cite{hou2020visual}. Table~\ref{table:obj_detector} illustrates better detected object boxes improves the performance largely. Meanwhile, we find ATL is apparently sensitive to worse boxes. Under worse object detector (\ie COCO detector), ATL does not improve the result. It might be because composing affordance features and object features from additional images results in poor generalization to worse boxes. When we transfer affordance representation to objects from a large number of additional images via composing novel HOI samples, we improve the scene generalization (\ie the model generalizes to novel scenes) of the affordance representation learning, while degrading the generalization to worse object boxes on HICO-DET test set. The object affordance recognition in Table~\ref{table:func_obj} illustrates the scene generalization of affordance and object representations. Noticeably, worse object detector largely hampers HOI detection in two-stage method. Thus, it is necessary to utilize better object detector for evaluating HOI detection, and ATL further improves HOI detection effectively with better object detector.

\begin{table}[tp]
\caption{Illustration of effect of domain shift on ATL between object images and HOI images on HOI-COCO dataset. Sub-COCO is a subset of COCO images that we randomly choose the same number of object instances to the objects of HICO-DET from COCO dataset.}

\label{table:hoi_domain}
\centering
\small
% \begin{center}

\begin{tabular}{@{}lcccc@{}}
\hline\hline
Method & Object images & Full & Rare & NonRare\cr

\hline\hline

% iCAN \cite{gao2018ican}
% Baseline & - & 22.86 &  6.87 &  35.27 \\
% VCL \cite{hou2020visual} & - & 23.53 & 8.29 & 35.36 \\
%   22.05 15.77 23.92
%  Ours     22.05 15.77 23.92
% Ours & HOI-COCO & 23.40 & 8.01 & 35.34 \\
ATL & HICO-DET & 24.21 & 9.52 & 35.61\\
ATL & Sub-COCO & {\bf 24.74} & {\bf 9.60} & {\bf 36.50}\\

\hline\hline
\end{tabular}

\end{table}

{\bf Domain difference}.
From the large performance gap between different object detectors in Table~\ref{table:obj_detector}, we find the HICO-DET dataset has a different domain to COCO. Table~\ref{table:hoi_domain} shows with the same number of object instances, COCO dataset improves the performance larger than HICO-DET dataset due to the domain difference on HOI-COCO. There is a similar trend in Table~\ref{table:sota_hico} and Table~\ref{table:sota_coco}. With the same COCO dataset, our method facilitates HOI detection on HOI-COCO dataset better than that on HICO-DET.

%
% \begin{table}[tp]
% \caption{Illustration of effect of the number of objects on HOI-COCO dataset.}
% \label{table:obj_nums}
% \centering
% \small
% % \begin{center}
%
% \begin{tabular}{@{}lcccc@{}}
% \hline
% The number of images & Full & Rare & NonRare\cr
%
% \hline\hline
%
% % iCAN \cite{gao2018ican}
% % Baseline & - & 22.86 &  6.87 &  35.27 \\
% % VCL \cite{hou2020visual} & - & 23.53 & 8.29 & 35.36 \\
% %   22.05 15.77 23.92
% %  Ours     22.05 15.77 23.92
% % Ours & HOI-COCO & 23.40 & 8.01 & 35.34 \\
% 1,000  & 24.14 & 8.93 & 35.94\\
% 2,500  & 24.21 & 9.52 & 35.61\\
% 5,000  & 24.76 & 8.93 & 37.05\\
% 10,000 &  25.07 & 10.19 & 36.62\\
% 20,000 &  24.77 & 9.45 & 36.67 \\
% % 30,000 &  24.49 & 9.05 & 36.48 \\
% Full & 24.84 & 9.79 & 36.51 \\
% \hline
% \end{tabular}
%
%
% \end{table}
%
% {\bf The size of object dataset}
% Changing object datasets from HOI-COCO to COCO 2017 in Table~\ref{table:sota_coco}, which effectively improves the performance, in fact increases the number of objects (\ie the size of object dataset). Then, we have a question How does HOI detection performance change as objects increase. We illustrate the effect of the number of objects in Table~\ref{table:obj_nums}. As the data set grows to 5,000 images, the performance improvement approaches saturation.
%

\begin{figure}
  \centering
  
  \includegraphics[width=0.4\textwidth]{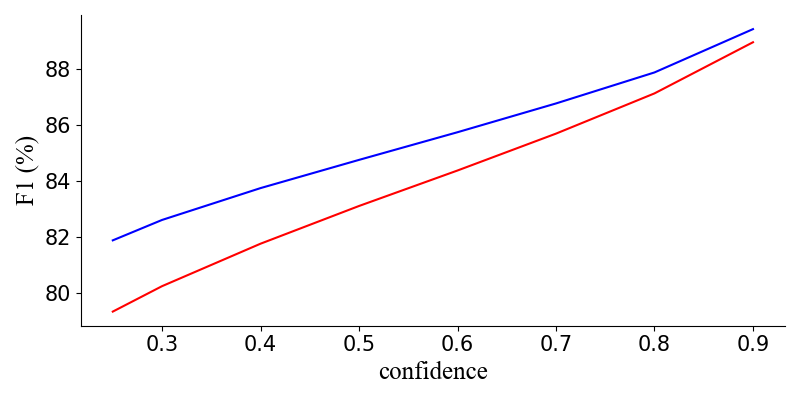}
  \caption{Comparison of object affordance recognition (F1) between ATL and the conversion from object detection results on HICO-DET. Confidence is the object detection confidence for choosing object boxes. Red is our method and Blue is the conversion from object detection results.}
  \label{fig:obj_func_det_f1}
\end{figure}

% \begin{table}[tp]
% \caption{Comparison of object functionality between the proposed method and the conversion from object detection results on HICO-DET. Confidence is the object detection confidence for choosing object boxes.}
% \label{table:func_detector}
% \centering
% \small
% % \begin{center}
%
% \begin{tabular}{@{}lcccc@{}}
% \hline
% Method & Precision & Recall & F1 & Confidence\cr
%
% \hline\hline
%
% functionality detection & 89.80 & 89.30 & 88.97  & 0.9 \\
% Ours & 91.25 & 89.25 &  89.44 & 0.9 \\
% functionality detection & 88.00 & 87.59 & 87.80  & 0.8 \\
% Ours & 89.88  & 87.75 &  88.80 & 0.8\\
% functionality detection &  87.31 & 86.94 & 86.43 & 0.75 \\
% Ours &89.37 & 87.22  & 87.33  & 0.75 \\
% functionality detection &  84.11 & 83.82  & 83.11 & 0.5 \\
% Ours & 86.99  & 84.85 &  85.91 &  0.5\\
% functionality detection &  80.41 & 80.38 & 79.33 & 0.25 \\
% Ours & 84.28 & 82.21 &  81.88 & 0.25 \\
%
% \hline
% \end{tabular}
%
% \end{table}

{\bf Affordance comparison with object detection results}. Our method can also be applied to detected boxes of an object detector. For a robust comparison, we directly compare ATL with the object affordance result converted from object detection results according to the object affordance annotation (\ie the ground truth affordances of an object category) on HICO-DET test set. Here we use the detected box of a COCO pretrained Faster-RCNN. We train our model on HOI-COCO dataset and COCO (2014) dataset, which has a same training set to COCO pretrained Faster-RCNN. Figure~\ref{fig:obj_func_det_f1} illustrates ATL achieves better affordance recognition results among different confidences. Meanwhile, ATL has better performance than object affordance detection when the confidence of detected box is lower.

\subsection{Qualitative Results}

\begin{figure}
  \centering
  \includegraphics[width=0.39\textwidth]{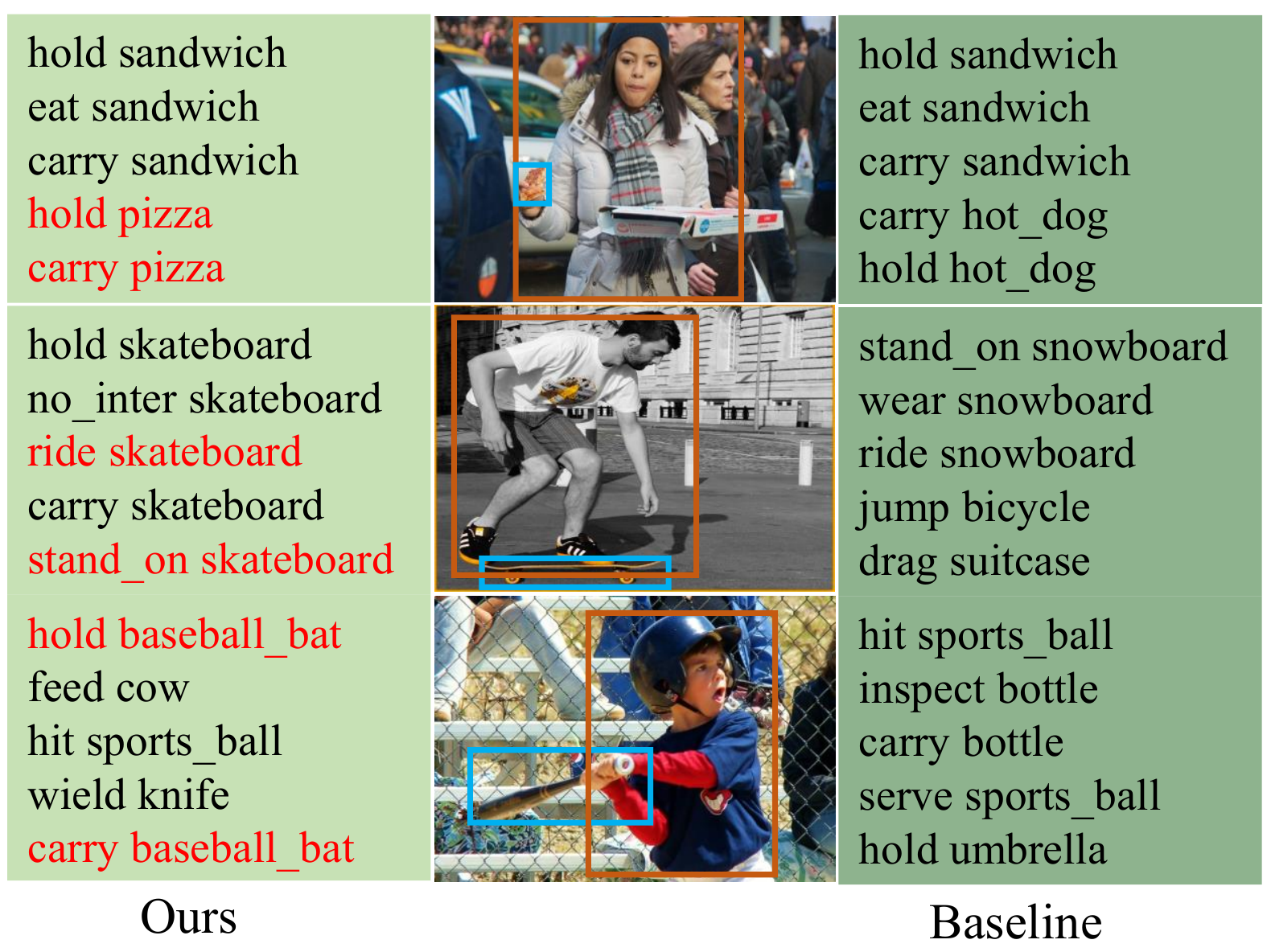}
  \caption{Illustration of unseen object zero-shot detection result (top 5) between the proposed method and Baseline. The correct results are highlighted in red.}
  \label{fig:zero_shot_fig_demo}
\end{figure}

We demonstrate the result of exploring unseen HOIs with novel objects in Figure~\ref{fig:zero_shot_fig_demo}. We find the baseline can not recognize the object at all, while the proposed method effectively detects the HOI with unseen objects.

% {\bf zero-shot images}

\section{Conclusion}
In this paper, we introduce a novel approach, affordance transfer learning or ATL, to transfer the affordance to novel objects via composing objects (from object images) and affordances (from HOI images) for HOI detection. ATL effectively facilitates HOI detection in long-tailed settings, especially for HOIs with novel objects. In addition, we devise a simple yet effective method to incorporate HOI detection model for object affordance recognition and ATL significantly improves the performance of the HOI detection model for object affordance recognition.

\noindent {\bf Acknowledgements} This work was supported in part by Australian Research Council Projects FL-170100117, DP-180103424, IH-180100002, and IC-190100031.

{\small
\bibliographystyle{ieee_fullname}
\bibliography{egbib}
}

\begin{appendix}

\section{Overview}
In this paper, we present an affordance transfer learning approach for Human-Object Interaction understanding and Object understanding in an unified way. More details and illustrations are introduced in appendix. We provide more examples between HOI and affordance in Section B. ATL for One-Stage HOI detection is illustrated in Section C. Section D contains more details. Section E shows more comparison between object affordance recognition and the ablation study of the number of verbs on affordance recognition. We illustrate the Non-COCO classes that we select from Object365 on section F. Section G provides additional affordance results (mAP) and additional illustration of recent HOI approaches. Lastly, we compare prior approaches (\ie VCL \cite{hou2020visual}, FCL \cite{hou2021fcl}) and ATL in detail.

\section{More Examples of HOI and Object Affordance}
Images labeled with HOI annotations simultaneously show the affordance of the objects. Therefore, we can not only learn to detect HOIs, but also learn to recognize the affordance of the objects. By combining the affordance representation with various kinds of its corresponding objects, we enable the HOI model to recognize the affordance of novel objects.

\begin{figure}
  \includegraphics[width=0.45\textwidth]{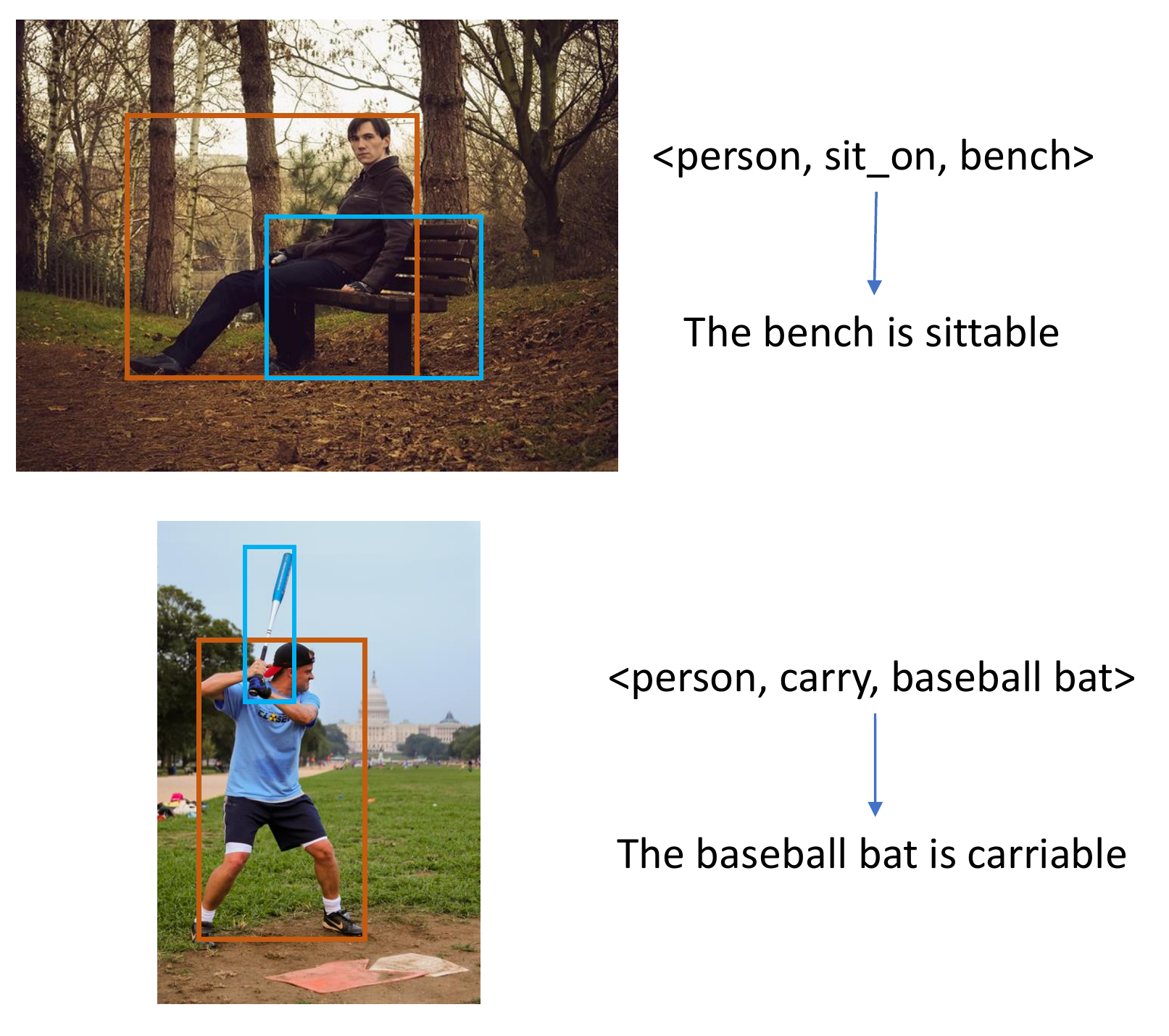}
  \caption{Examples about HOI and Affordance.}
  \label{fig:hoi_affordance}
\end{figure}

\begin{table*}[tp]
\caption{Additional Ablation study of object affordance recognition with HOI network among different number of object images on HICO-DET. Val2017 is the validation 2017 of COCO \cite{lin2014microsoft}. Subset of Object365 is the validation of Object365 \cite{Shao2020Objects365} with only COCO labels. Novel classes are selected from Object365 with non-COCO labels. Object means what object dataset we use. The content in parentheses indicates the number of images in each batch.}
\label{table:func_obj_img}
\centering
\small
% \begin{center}
\begin{tabular}{@{}lccccc|ccc|ccc|ccc@{}}
\hline\hline
\multirow{2}{*}{Method} & \multirow{2}{*}{HOI Data} & \multirow{2}{*}{Object}
& \multicolumn{3}{c}{Val2017 of COCO}&\multicolumn{3}{c}{Subset of Object365} &\multicolumn{3}{c}{HICO-DET} & \multicolumn{3}{c}{Novel classes}\cr\cline{4-15}
&&&Rec&Prec&F1&Rec&Prec&F1&Rec&Prec&F1&Rec&Prec&F1\\
\hline\hline

ATL (1) &HICO & COCO & 16.63 &	62.91&	24.73 & 12.47	&39.92&	17.45& 12.47&	52.45	&18.66 &7.30  &  18.44 &  9.73 \\
ATL (2) &HICO & COCO& {\bf 33.69}  & {\bf 79.54} & {\bf 44.32} & {\bf 28.25} & {\bf 63.56} & {\bf 35.24} & {\bf 30.27}&  {\bf 73.53}&  {\bf 40.31} & 12.41 &14.56 & 12.86 \\

ATL (3) &HICO & COCO&  27.36	& 78.21&	38.12 &  21.84&	53.57	&28.25 &   13.85	&57.42	&20.94 &  {\bf12.15}  &  {\bf26.07} & {\bf 15.56}	\\

\hline\hline
\end{tabular}

\end{table*}

\begin{table*}[tp]
\caption{Illustration of union verb representation on object affordance recognition. w/o union verb means we extract verb representation from the human box. Val2017 is the validation 2017 of COCO \cite{lin2014microsoft}. Object365 is the validation of Object365 \cite{Shao2020Objects365} with only COCO labels. Novel classes are selected from Object365 with non-COCO labels. Object means what object dataset we use. All results are reported by Mean average Precision (mAP)(\%).}
\label{table:func_obj_ap_ab}
\centering
\small
% \begin{center}
\begin{tabular}{@{}lccc|c|c|c@{}}
\hline\hline
Method & HOI Data & Object & Val2017 & Object365 & HICO-DET & Novel classes\cr\cline{4-7}
\hline\hline

Baseline &HICO&- &    19.71 &    17.86 &    23.18 &   6.80 \\
Baseline w/o union verb &HICO&- &    28.16 &    25.14 &    37.88 &   8.38 \\
\hline

ATL  &HICO & HICO&            {\bf 52.01}&        {\bf 50.94 }  &    {\bf  59.44} &     {\bf 15.64 }\\
ATL w/o union verb & HICO & HICO & 44.47 & 45.35 & 52.76 & 15.09 \\

% ATL &HICO & COCO&       {\bf 56.05 } &       40.83 &       57.41 & 8.52 \\

\hline\hline
\end{tabular}
\end{table*}

\section{Affordance Transfer Learning for One-Stage HOI detection}
Current HOI approaches mainly include one-stage methods \cite{liao2019ppdm, wang2020learning, wang2020discovering} and two-stage methods \cite{gao2018ican, hou2020visual}. In main paper, we simultaneously evaluate ATL on both one-stage method and two-stage method. We implement ATL based on the code of \cite{wang2020discovering}, which implement HOI detection based on Faster-RCNN \cite{ren2015faster}. In details, we use 2 object images and 4 HOI images for each batch with 2 GPUs. Here, we regard the concatenation of features extracted from union and human boxes with RoI Align separately as verb feature. We regard the feature extracted from object boxes as object feature. We compose novel HOIs from object features and verb features between HOI images and object images. Different from two-stage method, we also compose object features and verb features between HOI images (\ie VCL \cite{hou2020visual}). Meanwhile, in one-stage method, we directly predict 117 verbs, which are further combined with object detection result to construct HOI prediction. Besides, during optimization, we keep the object detection optimization for object images. Baseline is the model without compositional learning loss. Code is available at \url{https://github.com/zhihou7/HOI-CL-OneStage}.

\section{Supplementary Description}. In the Table 4 (object affordance recognition) in paper, we illustrate the affordance recognition of novel classes in zero-shot HOI detection on HICO-DET. All objects in HICO-DET, Val2017, Object365 with COCO classes are from the 12 novel classes (unseen objects). In the Novel Classes category, those objects are still Non-COCO classes.  Besides, we can use both COCO images and HOI images as object images in our experiments. But we do not find any improvement on HICO-DET. Thus in Table~\ref{table:sota_hico}, we do not include the result when we use two datasets as object images. It might be because there are nearly 900,000 object instances in COCO while HICO has only around 100, 000 object instances. For novel object zero-shot, there are too much many composite HOIs for seen HOIs, we thus remove some COCO object images for balancing the data. Besides, similar to \cite{hou2021fcl}, we also fuse HOI detection result to object detection result to improve the baseline. In addition, in our experiment, the baseline (ATL with only HICO data as object images) converges faster because the number of training data is much less than ATL (COCO). For simplicity, we directly fine-tune ATL (HICO) and ATL (COCO) models.

\begin{table*}[tp]
\caption{The effect of different number of verbs in affordance feature bank. Mean average Precision (mAP) (\%) is reported. Dataset means the evaluation object dataset. HICO-DET means the test set of HICO-DET. Val2017 means the validation set of COCO2017. }
\label{table:afford_insts}
\centering
\small
% \begin{center}

\begin{tabular}{@{}llccccccc@{}}
\hline\hline
\#$M$ & Dataset & 1 & 5 & 10 & 20 & 40 & 80 & 100 \cr
\hline
Baseline & Val2017 & 13.39 & 15.90 &  17.69 & 18.74 & 19.25 & 19.67 & {\bf 19.71} \\
ATL (COCO) &Val2017 & 52.98 & 53.74 & 55.40 & 55.19 & 54.88 & 55.77 & {\bf 56.05} \\
Baseline & HICO-DET & 14.77 & 18.30 & 20.22 & 21.70 & 22.21 & 23.00 & {\bf 23.18} \\
ATL (COCO) & HICO-DET & 56.04 & 58.03 & {\bf 59.14} &  57.84&  56.61&  57.23 & 57.41 \\

\hline\hline
\end{tabular}
\end{table*}

\section{Additional Ablation Study}

{\bf The effect of different number of object images on affordance recognition}. Table~\ref{table:func_obj_img} illustrates the comparison of object affordance recognition among different number of object images in the minibatch on HICO-DET dataset. We find ATL with two images in each batch apparently improves the performance of object affordance recognition compared to one image and three images among COCO categories. Moreover, we find with more object images in each batch, ATL further improves the affordance recognition performance on Non-COCO classes. This means with multiple object images, ATL has better generalization of affordance recognition to novel classes.

{\bf Union verb representation}. Following~\cite{hou2020visual}, we extract verb representation from the union box in our experiments. The effect of union verb representation~\cite{hou2020visual} on HOI detection is relatively small. However, we notice the union verb representation has a significant effect on object affordance recognition. Table~\ref{table:func_obj_ap_ab} illustrates the results of ATL on affordance recognition drops by over 5\% on COCO, Object365 and HICO-DET dataset when extracting verb representation from human box. However, for the baseline model, the performance of affordance recognition improves when we extract verb representation from human box. This might because the union box also includes some noise information (the region out of human and object box), while the compositional approach facilitates the verb representation learning from union box, \ie extract useful information for verb representation from the union box. Noticeably, the performances of the two verb representations on HOI detection are not very different. {\bf We think object affordance recognition also provides a benchmark for the evaluate of verb (action) representation learning.}

{\bf The effect of the number of verbs on affordance recognition.} In affordance recognition, we randomly choose $M$ instances for those affordances with more than $M$ instances in dataset and all instances for other affordances. We ablate $M$ in Table~\ref{table:afford_insts} under the ATL model with COCO objects and our baseline. The baseline is the model without compositional learning. Besides, when we use different $M$, we also update $S_{i}$. If we keep $S_{i}$ same as the number when $M=100$, all results will be very small when $M<100$.

Table~\ref{table:afford_insts} shows the number goes stable after $20$. This means we do not need to store a large number of templates of affordance representation.

\section{Non-COCO classes}
For evaluating ATL on affordance recognition of unseen classes, we manually select 12 non-coco classes from object365: glove, microphone, american football, strawberry, flashlight, tape, baozi, durian, boots, ship, flower, basketball. The actions that we can act on those objects (\ie affordance) on HOI-COCO and HICO-DET are list on Table~\ref{table:affordance_illu1} and Table~\ref{table:affordance_illu2} respectively.

% In paper, some affordance labels of the Non-COCO classes in Object365 on HICO-DET are incorrect. Thus, all results of novel classes in Table 5 in paper are low. We revise the annotations to Table~\ref{table:affordance_illu2}. The revised result of HICO-DET on novel classes is illustrated in Table~\ref{table:func_obj_novel}. We can find similar trend in Table~\ref{table:func_obj_novel}. We will revise the about novel classes on HICO-DET in Table 5 in paper. 

\begin{figure}
  \centering
  \includegraphics[width=0.5\textwidth]{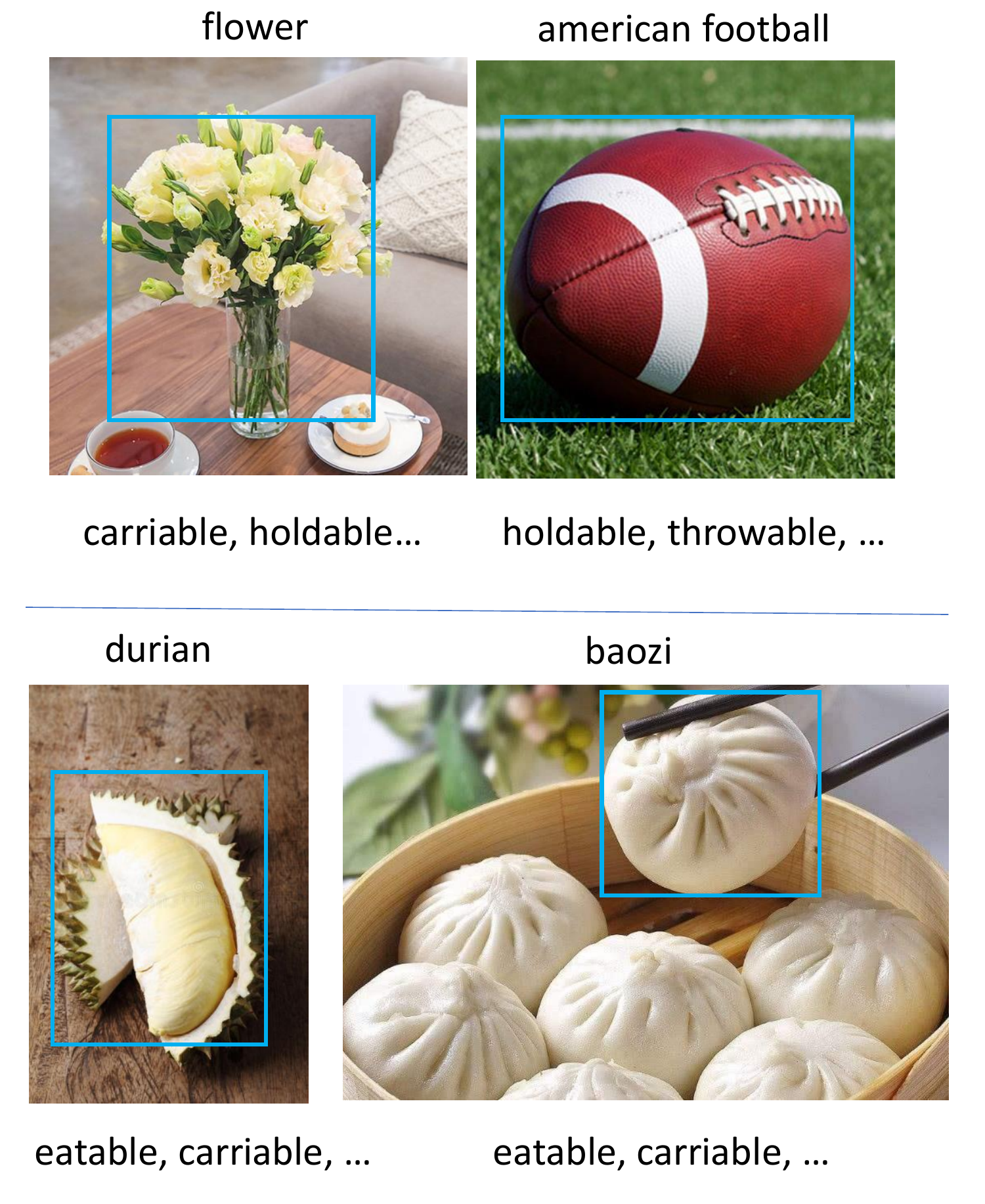}
  \caption{Examples of Non-COCO classes and its affordances.}
  \label{fig:eg_affordance}
\end{figure}

We further provide some visual examples of the Non-COCO classes in Figure~\ref{fig:eg_affordance}. ATL can recognize the affordance of those objects without being interacted by combining the affordance representation and those object features.

\begin{table}[tp]
\caption{Affordances of Non-COCO classes from Object365 on HOI-COCO.}
\label{table:affordance_illu1}
\centering
\small
% \begin{center}
\begin{tabular}{@{}lc@{}}
\hline\hline
name & verbs/affordances\\
\hline\hline
glove & carry, throw, hold\\
microphone & talk\_on\_phone, carry, throw, look, hold\\
american football & kick, carry, throw, look, hit, hold\\
strawberry & cut, eat, carry, throw, hold\\
flashlight & carry, throw, hold \\
tape & carry, throw, hold\\
baozi & eat, carry, look, hold\\
durian & eat, carry, hold\\
boots & carry, hold \\
ship & ride, sit, lay, look\\
flower & look, hold \\
basketball & throw, hold\\

\hline\hline
\end{tabular}

\end{table}

\begin{table}[tp]
\caption{Affordances of Non-COCO classes from Object365 on HICO-DET.}
\label{table:affordance_illu2}
\centering
\small
% \begin{center}
\begin{tabular}{@{}lc@{}}
\hline\hline
name & verbs/affordances\\
\hline\hline

glove & buy, carry, hold, lift, pick\_up, wear \\
microphone & carry, hold, lift, pick\_up \\
american football & block, carry, catch, hold, kick, lift, pick\_up, throw \\
strawberry & buy, eat, hold, lift, move \\
flashlight & buy, hold, lift, pick\_up  \\
tape & buy, hold, lift, pick\_up \\
baozi &  buy, eat, hold, lift, pick\_up\\
durian &  buy, hold, lift, pick\_up\\
boots & buy, hold, lift, pick\_up, wear \\
ship & adjust, board \\
flower & buy, hold, hose, lift, pick\_up  \\
basketball & block, hold, kick, lift, pick\_up, throw \\
\hline\hline
\end{tabular}

\end{table}

\section{Additional Results and Comparision}
We find the metrics (Recall, Precision, F1) the paper (first version) uses is not much robust. F1 is sensitive to the confidence. Thus, we further evaluate the affordance recognition in Table~\ref{table:func_obj_ap} by Mean average Precision (mAP) (\%). Table~\ref{table:func_obj_ap} shows the compositional learning approach consistently improves the baseline among all categories.

Due to the limitation of space in main paper. Other recent HOI detection methods are provided in Table~\ref{table:sota_hico_add}.

\section{Discussion with Prior Approaches}
ATL extends VCL \cite{hou2020visual} by composing verbs and objects from object detection datasets which do not have HOI annotations. ATL presents a way to explore a broader source of data for HOI detection. Meanwhile, ATL finds that the HOI network trained with compositional learning can be simultaneously applied to affordance recognition. Meanwhile, ATL shows with more data, ATL can improve the generalization of affordance recognition on new dataset.

\begin{table*}[tp]
\caption{Comparison of object affordance recognition with HOI network among different datasets. Val2017 is the validation 2017 of COCO \cite{lin2014microsoft}. Object365 is the validation of Object365 \cite{Shao2020Objects365} with only COCO labels. Novel classes are selected from Object365 with non-COCO labels. Object means what object dataset we use. ATL$^{ZS}$ means novel object zero-shot HOI detection model in Table 3 on HICO-DET. For ATL$^{ZS}$, we show the results of the 12 classes of novel objects in Val2017, Subset of Object365 and HICO-DET. All results are reported by Mean average Precision (mAP)(\%).}
\label{table:func_obj_ap}
\centering
\small
% \begin{center}
\begin{tabular}{@{}lccc|c|c|c@{}}
\hline\hline
Method & HOI Data & Object & Val2017 & Object365 & HICO-DET & Novel classes\cr\cline{4-7}
\hline\hline
Baseline & HOI & - &      31.91 &  26.16 &     44.00 &      14.27  \\
% VCL \cite{hou2020visual} & 21.96 & 19.56 & 12.79 & 21.58 & 22.05 & 15.77 & 23.92  & 23.92 & 23.92\\ 9.62
FCL \cite{hou2021fcl} & HOI & - &      41.89 &    32.20 &       55.95&   18.84 \\
VCL \cite{hou2020visual} &  HOI  & HOI&        76.43   &       69.04 &  86.89 &        32.36\\
ATL & HOI  & HOI &   76.52 &       69.27 &     87.20 &   34.20 \\
ATL &  HOI & COCO &     {\bf  90.84} & {\bf  85.83} & {\bf 92.79} &   {\bf 36.28}\\
\hline
Baseline &HICO&- &    19.71 &    17.86 &    23.18 &   6.80 \\
FCL \cite{hou2021fcl} & HICO & - &       25.11    &     25.21 &         37.32 &    6.80 \\
VCL \cite{hou2020visual} &HICO & HICO &     36.74 &  35.73   &       43.15 &  12.05 \\
ATL  &HICO & HICO&            52.01 &        {\bf 50.94 }  &    {\bf  59.44} &     {\bf 15.64 }\\
ATL &HICO & COCO&       {\bf 56.05 } &       40.83 &       57.41 & 8.52 \\
\hline
ATL$^{ZS}$   &HICO & HICO&      24.21 &       20.88 &     28.56 &       12.26 \\
ATL$^{ZS}$ &HICO & COCO &    {\bf 35.55} &   {\bf 31.77} &   {\bf  39.45} &    {\bf13.25} \\
\hline\hline
\end{tabular}
\end{table*}

% \begin{table}[tp]
% \caption{Results of Object Affordance Recognition on Novel classes after correcting the affordance labels of novel classes. Novel classes are selected from Object365 with non-COCO labels. Object means what object dataset we use.}
% \label{table:func_obj_novel}
% \centering
% \small
% % \begin{center}
% \begin{tabular}{@{}lccccc@{}}
% \hline\hline
% Method & HOI Data & Object& Rec&Prec&F1\\
% \hline\hline

% \hline
% % Baseline &HICO&- & 4.14&	17.14&	6.47 \\
% % VCL \cite{hou2020visual} &HICO & HICO &  2.01&	6.75&	2.98 \\
% % ATL  &HICO & HICO& {\bf 8.63} &	{\bf 20.67} &	{\bf 11.58} \\
% % ATL &HICO & COCO& 7.62&	19.75& 10.42 \\

% Baseline &HICO&- & 8.12&	15.87&	8.78 \\
% VCL \cite{hou2020visual} &HICO &HICO &7.81&	22.63&	11.02\\
% ATL  &HICO & HICO& {\bf 12.78}&	{\bf 28.8}& {\bf	16.78} \\
% ATL &HICO & COCO& 12.41&	14.56&	12.86 \\

% \hline
% ATL$^{ZS}$ &HICO & HICO&   5.02 &   11.63 &  6.79\\
% ATL$^{ZS}$   &HICO & COCO& {\bf 14.00} & {\bf 28.60} & {\bf  18.07} \\
% \hline

% % ATL$^{ZS}$   &HICO & HICO& 1.64	&4.43	&2.35 \\

% % ATL$^{ZS}$ &HICO & COCO&{\bf 7.58}&{\bf	21.6}&{\bf	10.93} \\

% \hline\hline
% \end{tabular}

% \end{table}

\setlength{\tabcolsep}{4pt}
\begin{table}[tp]
\centering
\caption{Comparison of Zero Shot Detection results of between FCL \cite{hou2021fcl} and ATL. NO means novel object HOI detection. * means we only use the boxes of the detection results.}
\label{tab:zero_shot_fcl}
\small
\begin{tabular}{@{}lcccc@{}}
\hline\hline
Method & Type & Unseen & Seen & Full \cr
\hline\hline
FCL \cite{hou2021fcl} & NO & 15.38 & 21.30 & 20.32 \\
ATL (COCO) &  NO & 15.11 &  {\bf 21.54}  & {\bf 20.47} \\
\hline
FCL \cite{hou2021fcl} & NO & 0.00 & 13.71 & 11.43\\
ATL (COCO)* &NO & {\bf 5.05} & {\bf 14.69} & {\bf 13.08} \\
\hline\hline
\end{tabular}
\end{table}

\setlength{\tabcolsep}{4pt}
\begin{table}[tp]
\centering
\caption{Evaluation of the complementary between ATL and FCL. We use the released model of FCL \cite{hou2021fcl}.}
\label{tab:compare_complementary}
\small
\begin{tabular}{@{}lcccc@{}}
\hline\hline
Method & Full & Rare & Non-Rare \cr
\hline\hline
FCL \cite{hou2021fcl} & 24.68 &20.03 &26.07 \\
ATL (COCO) & 24.50  &  18.53 & 26.28 \\
FCL + ATL & 25.63 &  21.18  & 26.95 \\
\hline\hline
\end{tabular}
\end{table}

\begin{table}[tp]
\caption{Additional Illustration of recent HOI detection approaches.}
\label{table:sota_hico_add}
\centering
% \begin{center}
% \small
\resizebox{0.95\linewidth}{!}{
\begin{tabular}{@{}lcccccc@{}}
\hline
\multirow{2}{*}{Method} &
\multicolumn{3}{c}{Default}&\multicolumn{3}{c}{Known Object}\cr\cline{2-7}
% \cmidrule(lr){2-4} \cmidrule(lr){5-7}
&Full&Rare&NonRare&Full&Rare&NonRare\cr
% \begin{tabular}{@{}llllllllll@{}}
% \multirow{Method}& Feature Backbone &
% \multicolumn{3}{c}{Default}&\multicolumn{3}{c}{Known Object}\cr
% & Full & Rare & NonRare & Full & Rare & NonRare\\
% \midrule
\hline\hline
% Shen \etal \cite{shen2018scaling} & VGG-19 & 6.46 & 4.24 & 7.12 & - & - & -  \\
% HO-RCNN \cite{chao2018learning}& CaffeNet & 7.81 & 5.37 & 8.54 & 10.41 & 8.94 & 10.85  \\
% InteractNet \cite{gkioxari2018detecting} & ResNet-50 & 9.94 & 7.16 & 10.77 & - &- & -  \\
% GPNN \cite{qi2018learning}  & ResNet-152 & 13.11 & 9.34 & 14.23 & - & - & -  \\
% iCAN \cite{gao2018ican} & COCO &14.84 & 10.45 & 16.15 & 16.26 & 11.33 & 17.73  \\
% Xu \etal \cite{xu2019learning}& COCO & 14.70 & 13.26 & 15.13 & - & - & -  \\
% Li \etal\cite{li2018transferable} & COCO & 17.03 & 13.42 & 18.11 & 19.17 & 15.51 & 20.26 \\
% Wang \etal \cite{wang2019deep} & COCO & 16.24 & 11.16 & 17.75 & 17.73 & 12.78 & 19.21 \\
% Gupta \etal \cite{gupta2019no} & COCO & 17.18 & 12.17 & 18.68 & - & - & -  \\
% Zhou \etal \cite{zhou2019relation} & COCO & 17.35 & 12.78 & 18.71 & - & - & - \\
% % Zhou \etal \cite{zhou2019relation}  & 17.35 & 12.78 & 18.71 & - & - & - & 47.5 \\
% PMFNet \cite{wan2019pose} & COCO & 17.46 & 15.65 & 18.00 & 20.34 & 17.47 & 21.20 \\
% Peyre \etal \cite{preye2019detecting} & COCO & 19.40 & 14.63 & 20.87 & - & - & -  \\
FG \cite{bansal2019detecting}  & 21.96 & 16.43 & 23.62 & - & - & - \\
VSGNet \cite{ulutan2020vsgnet} & 19.80 & 16.05 & 20.91 & - & - & - \\
DJ-RN \cite{li2020detailed} & 21.34 & 18.53 & 22.18 & 23.69 & 20.64 & 24.60 \\
% \hline
IP-Net \cite{wang2020learning} & 19.56 & 12.79 & 21.58 & 22.05 & 15.77 & 23.92 \\
PPDM \cite{liao2019ppdm} &  21.73 & 13.78 &24.10 &24.58 &16.65 &26.84 \\
Kim \etal \cite{kim2020uniondet} & 17.58 & 11.72 & 19.33 & 19.76 & 14.68 & 21.27 \\
ACP \cite{kim2020detecting} & 20.59 & 15.92 & 21.98 & - & - & - \\
PD-Net \cite{zhong2020polysemy} & 20.81 & 15.90 & 22.28 & 24.78 & 18.88& 26.54\\
FCMNet \cite{liuamplifying} & 20.41& 17.34& 21.56 &22.04& 18.97 &23.12 \\
VCL \cite{hou2020visual} & 23.63 & 17.21 & 25.55 & 25.98 & 19.12 & 28.03 \\
DRG \cite{gao2020drg} & 24.53 & 19.47 & 26.04 & 27.98 & 23.11 & 29.43 \\
%   22.05 15.77 23.92
%  Ours     22.05 15.77 23.92

\hline
% 0.1855  0.5302    0.1391  0.5418     0.1993
% Ours & Resnet-50 & 23.43 & 18.23 & 24.99 & 25.33 & 19.82 & {\bf 26.97} \\
ATL (COCO) $^{VCL}$ & 24.50 & 18.53 & 26.28 &   27.23  &  21.27  &   29.00  \\

ATL (COCO) $^{DRG}$ &   28.53   &    21.64   &   30.59  &     31.18    &   24.15    &   33.29  \\

\hline
\end{tabular}
}
% \end{center}
% \caption{Comparisons with the state-of-the-art approaches on HICO-DET dataset\cite{chao2018learning}. Xu \etal \cite{xu2019learning} and Peyre \etal \cite{peyre2018detecting} utilize language knowledge. Mean average Precision (mAP) (\%) for the
% default setting and known object setting is reported where higher values indicates better performance. The best scores are marked in {\bf bold}.}
\end{table}

Prior to ATL, Fabricated Compositional Learning \cite{hou2021fcl} was presented to fabricated objects to ease the open long-tailed issue for HOI detection. FCL \cite{hou2021fcl} inspires our to compose novel HOIs from verb features from HOI images and object features from external object datasets. Compared to VCL \cite{hou2020visual} and ATL \cite{hou2020visual}, FCL \cite{hou2020visual} is more flexible to generate balanced objects for each verb, and thus achieves better performance on some zero-shot settings. However, FCL also has some limitations. Although FCL achieves similar even better performance to ATL in HOI detection, Table~\ref{table:func_obj_ap} shows the model of FCL in fact is unable to recognize affordance. Besides, Table~\ref{tab:zero_shot_fcl} further shows although FCL \cite{hou2020visual} achieves also good results on Novel Object HOI detection with a generic object detector, the results of FCL~\cite{hou2020visual} on Unseen category drop to 0 without generic object detector. 

We further illustrates the complementary between FCL and VCL in Table~\ref{tab:compare_complementary}. Here, we fuse the prediction results of the two model to evaluate the complementary. We find this can largely improves the result.

\end{appendix}
\end{document}